\documentclass[letterpaper, 10 pt, conference]{ieeeconf}  

\IEEEoverridecommandlockouts                              
\usepackage{graphicx} 
\usepackage{amsmath} 
\usepackage{amssymb}  
\usepackage{xfrac}
\usepackage{gensymb}
\usepackage{subfig}
\let\labelindent\relax
\usepackage[shortlabels]{enumitem}

\bstctlcite{IEEEexample:BSTcontrol}

\title{\LARGE \bf Dynamics and Aerial Attitude Control for Rapid Emergency Deployment of the Agile Ground Robot AGRO}
\author{Daniel J. Gonzalez$^{1*}$, \IEEEmembership{Member,~IEEE}, Mark C. Lesak$^{2}$, \IEEEmembership{Member,~IEEE}, Andres H. Rodriguez$^{1}$,\\ Joseph A. Cymerman$^{3}$, and Christopher M. Korpela$^{1}$, \IEEEmembership{Senior Member,~IEEE}
\thanks{$^{*}$Corresponding author email: {\tt\small dgonzrobotics@gmail.com}}
\thanks{$^{1}$Robotics Research Center, Department of Electrical Engineering and Computer Science. Email: {\tt\small \{daniel.gonzalez, andres.rodriguez, christopher.korpela\}@westpoint.edu}.}
\thanks{$^{2}$Army Cyber Institute. Email: {\tt\small mark.lesak@westpoint.edu}.}
\thanks{$^{3}$Department of Civil and Mechanical Engineering. Email: {\tt\small joseph.cymerman@westpoint.edu}.}
\thanks{$^{1, 2, 3}$United States Military Academy, West Point, NY 10996, USA.}
}

\begin{document}
\bstctlcite{IEEEexample:BSTcontrol}

\maketitle
\thispagestyle{empty}
\pagestyle{empty}

\begin{abstract}
In this work we present a Four-Wheeled Independent Drive and Steering (4WIDS) robot named AGRO and a method of controlling its orientation while airborne using wheel reaction torques. This is the first documented use of independently steerable wheels to both drive on the ground and achieve aerial attitude control when thrown. Inspired by a cat's self-righting reflex, this capability was developed to allow emergency response personnel to rapidly deploy AGRO by throwing it over walls and fences or through windows without the risk of it landing upside down. It also allows AGRO to drive off of ledges and ensure it lands on all four wheels. We have demonstrated a successful thrown deployment of AGRO.

A novel parametrization and singularity analysis of 4WIDS kinematics reveals independent yaw authority with simultaneous adjustment of the ratio between roll and pitch authority. Simple PD controllers allow for stabilization of roll, pitch, and yaw. These controllers were tested in a simulation using derived dynamic equations of motion, then implemented on the AGRO prototype. An experiment comparing a controlled and non-controlled fall was conducted in which AGRO was dropped from a height of 0.85 m with an initial roll and pitch angle of 16 degrees and -23 degrees respectively. With the controller enabled, AGRO can use the reaction torque from its wheels to stabilize its orientation within 402 milliseconds. 

Keywords: Wheeled Robots, Dynamics, Motion Control
\end{abstract}


\section{Introduction}\label{Intro}
The ubiquitous and rapid deployment of robots alongside or instead of humans during emergency and disaster situations has long been a goal of the global robotics research community \cite{Murphy2012}. The DARPA Robotics Challenge catalyzed the research and development of humanoid and human-sized rescue robots and the algorithms to control and operate them reliably \cite{Atkeson2015}, but disaster robots that move with wheels and treads see the most adoption, with the successful deployment of the Colossus robot to help combat the 2019 Notr\'e Dame fire being a recent highlight \cite{Peskoe-Yang2019}. Wheels or treads provide robots the means of quickly, efficiently, and reliably traversing flat land and driving over some small obstacles. Most wheeled robots, however, are designed with either skid-steer \cite{Yi2009} or Ackerman \cite{Weinstein2010} steering geometries that limit their mobility, and are easily blocked by stairs, curbs, ditches, and other obstacles depending on their design. These robots can tip onto their side or back when pushed beyond their physical limits, requiring operator intervention if not equipped with a self-righting mechanism \cite{Kessens2012}. On the other hand, legged robots can move omnidirectionally and traverse such obstacles either by stepping over them or jumping and landing, but commercial offerings still lack the speed, efficiency, and reliability of their wheeled counterparts. 

\begin{figure}[t]
	\centering
	\includegraphics[width=1\linewidth]{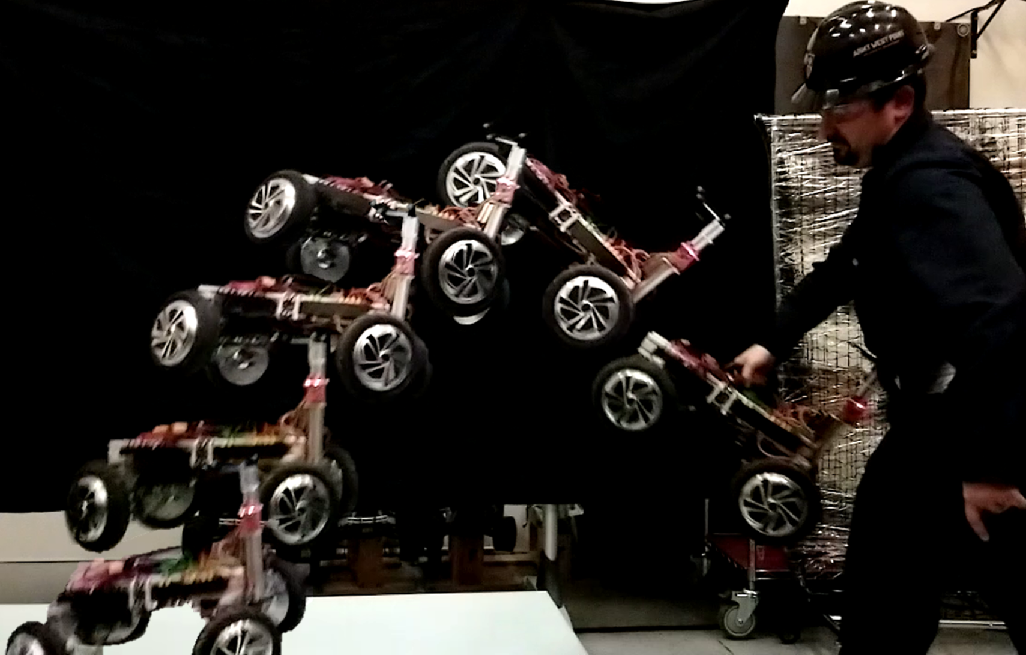}
	\caption{Composite video stills of AGRO being thrown, stabilizing its attitude mid-air, and landing on all four wheels. \textbf{A video of this demonstration is available in the supplemental materials for this work.}}
	\label{Throw}
\end{figure}
AGRO is a novel Agile Ground RObot that aims to combine the best attributes of both legged and wheeled platforms in its design to be highly maneuverable and rapidly deployable in emergency situations (See Fig. \ref{Throw}). 
The AGRO prototype has a Four-Wheel Independent Drive and Steering (4WIDS) architecture similar to that of ODV9 \cite{Mori2002} and AZIMUT \cite{Michaud2003} that enables it to maneuver quasi-omnidirectionally on the ground.

This wheel architecture also allows AGRO the novel capability of controlling its orientation in the air and landing on its ``feet'' like a cat.
A cat twists its body and legs while conserving angular momentum to reorient itself (first documented by Marey in 1894 \cite{Marey1894} and then demonstrated on a twisting robot \cite{Mather2009}). AGRO instead uses the reaction torques from its in-wheel hub motors to control its base orientation mid-air, land upright, and distribute the force of impact evenly to all four wheels. This is an important capability because it allows AGRO to be rapidly and reliably deployed by emergency response personnel by being thrown over walls and fences, or through windows without the risk of landing upside-down.

While this is the first documented use of multiple independently steerable wheels to both drive on the ground and achieve aerial attitude control for a throwable robot, the use of reaction torque to control orientation has been studied extensively, perhaps most for the control of spacecraft and satellite orientation using flywheels \cite{Rui2000}. 
A variety of legged robots have been demonstrated controlling their balance and orientation using a tail \cite{Briggs2012} \cite{Brill2015}, and an independent flywheel \cite{Kolvenbach2020} \cite{Yim2018}.
In the sport of motocross, drivers manually control the orientation of their motorcycles to perform tricks and safely land from jumps by accelerating or braking their rear wheel \cite{Giles1996}. 

In this work, we enable the rapid deployment and aerial attitude control of robots with independently steerable and driven wheels by throwing. We analyze and demonstrate the use of wheel reaction torque for dynamic attitude control of a class of 4WIDS robots and implement a simple pitch and roll controller on the AGRO prototype. 

The Agile Ground Robot concept and hardware design are discussed in Section \ref{AGRO Overview}. In Section \ref{Analysis} the affect of kinematics on attitude control authority are analyzed and the dynamic equations of motion of the airborne robot are derived. A control strategy for stabilizing aerial attitude is described and simulated in Section \ref{Control}. This control strategy is demonstrated on the robot prototype in Section \ref{Experiment} and compared to uncontrolled motion. Section \ref{Conclusion} provides a conclusion and outlines future work to be conducted.

\section{AGRO the Agile Ground RObot}\label{AGRO Overview}
\subsection{Agile Emergency Responder Concept}\label{Concept}
AGRO was designed out of a need for agile, easy-to-use, reliable, and relatively inexpensive ground robot platforms that could maneuver both indoors and in outdoor urban and off-road environments while carrying out teleoperated or autonomous inspection and response missions. For example, such a robot could be equipped with a mapping and radiation detection sensor and driven outside a compound or inside a contaminated building to generate a map with radiation hotspots identified. This map can then provide emergency responders the information they need to plan safe movements and reduce their risk of injury. A similar system could be used to respond to a chemical spill or a fire. 

If entryways to the disaster location are locked, blocked by debris, or otherwise barred, AGRO can be deployed by being thrown. For example, from a firefighting ladder truck, AGRO can be thrown through a window into a burning building, or over a wall into a nuclear facility. The ability to reliably and repeatably land on its wheels is critical for the continuation of the mission. 


\subsection{Robot Hardware Design Overview}\label{Prototype}
\begin{figure}[t]
	\centering
	\includegraphics[width=1\linewidth]{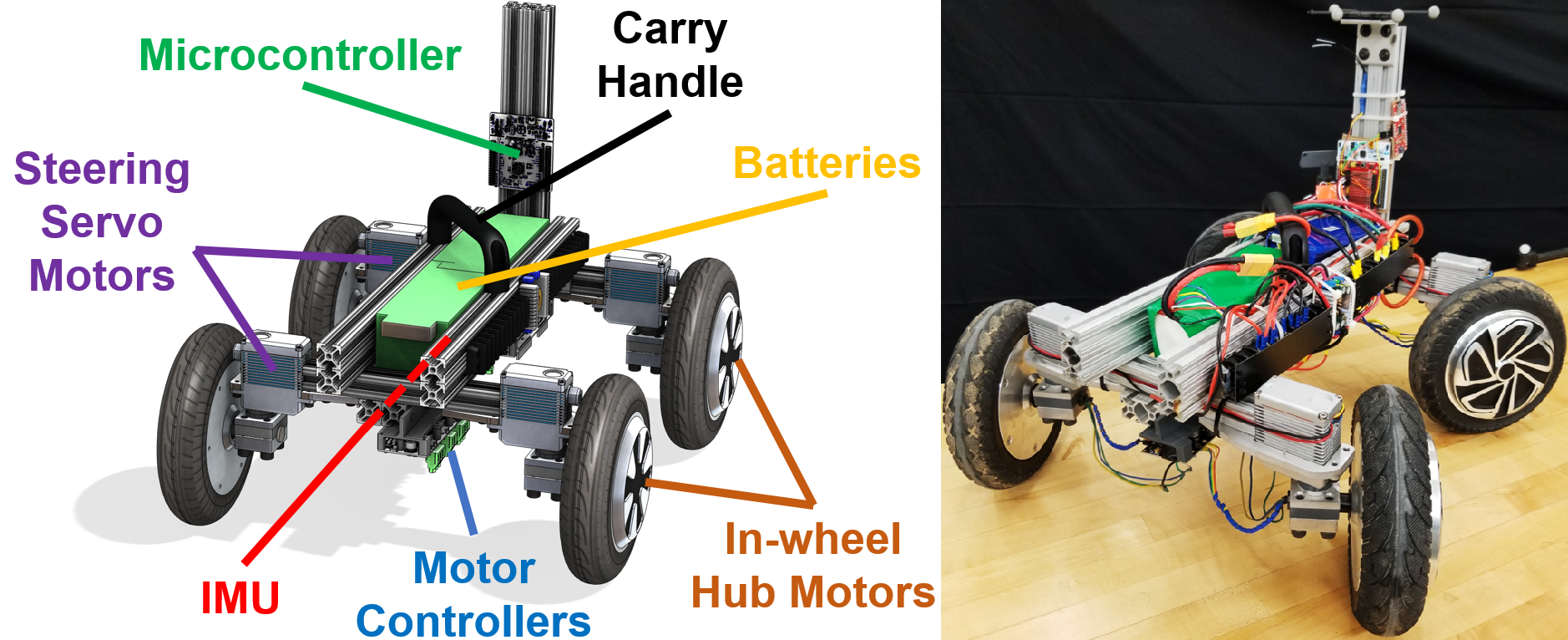}
	\caption{AGRO \iffalse MK I \fi Solid Model, Design Overview, and Prototype.}
	\label{DesignFig}
\end{figure}
The AGRO prototype (See Fig. \ref{DesignFig}) incorporates 4 steering and drive modules. Each module consists of one ``Hoverboard'' in-wheel hub motor for drive and a Hitec HS-1000SGT servo motor for steering attached to each wheel through custom mounting brackets. The front pair and rear pair of motors are each driven by an ODrive two-channel brushless motor controller. Two Turnigy 6s 16Ah Lithium Polymer batteries connected in series supply 44.4V for motor drive. A DROK DCDC converter provides 12V at 15A peak to power the four servo motors. The drive system allows AGRO to achieve a top speed of 11.4 m/s (40.9 km/h or 25.4 mph), enabling rapid movement from building-to-building. 

The structure is made from 80/20 t-slotted aluminum extrusion and custom waterjet aluminum brackets. The batteries are cradled in the center of the chassis for close proximity to the center of mass. A Redshift Labs UM7 Orientation Sensor is mounted below the main base, also near the center of mass, and measures the base orientation in Euler angles, the base angular velocity, and the base accelerations. The ODrives are mounted to the chassis using 3D printed components and are located between each pair of wheels for ease of motor cable management. A carry handle is mounted on the chassis to facilitate deployment.

On the ``tail pillar'' sits an Mbed Nucleo F446RE microcontroller which performs all control calculations for this system. A SparkFun Logomatic v2 microSD card read/write module is connected to the Mbed for internal logging. An FrSky R-XSR RC radio receiver is also mounted on this ``tail pillar'' and accepts commands from an FrSky Taranis QX7 transmitter. A Hella Master Power Switch is used to switch main battery power. Power wiring is facilitated by terminal blocks mounted on the left side of the chassis. 
The entire prototype robot has a mass of 23.59 kg (52 lbs).

\section{Kinematic and Dynamic Analysis}\label{Analysis}
New to this application is the aerial phase of motion in which both the motor wheel reaction torques and the reaction forces of steering can affect the orientation of the robot base. We will investigate how the kinematic configuration affects aerial attitude control authority, then derive dynamic equations of motion. 

\begin{figure}[h]
	\centering
	\includegraphics[width=0.875\linewidth]{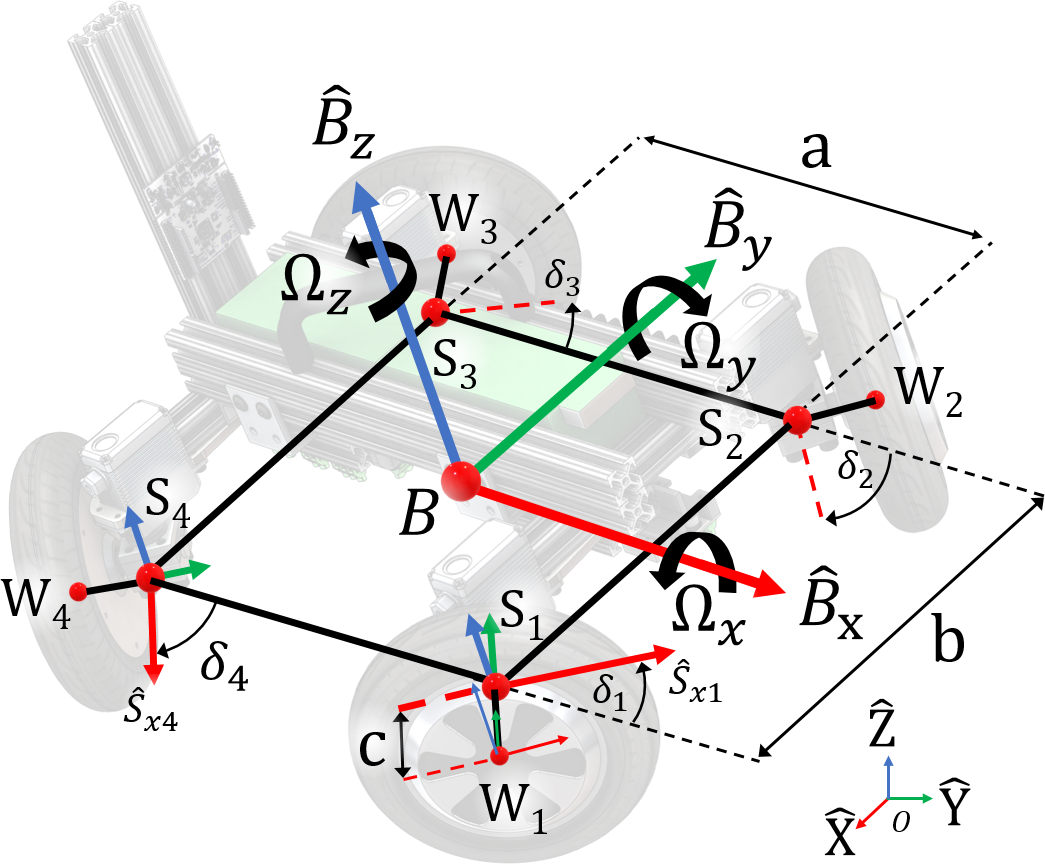}
	
	\caption{3D Airborne Kinematic Parameters.}
	\label{Kinematic Params 3D}
\end{figure}
The model consists of a rigid body $B$ representing the base of the robot and its fixed components (See Fig. \ref{Kinematic Params 3D}). Point $B$ represents the fixed center of mass of the base. Four wheels, also modelled as rigid bodies, connect to $B$ via massless linkages at steering points $S_i$, which fix each respective wheel to the base. Points $W_i$ account for the mass centers of each wheel. All mass centers are assumed to be coplanar. A right-handed, orthogonal, and unitary vector basis $\hat{B}_{xyz}$ is fixed to base $B$ and the vector bases $\hat{W}_{xyz,i}$ are fixed to each respective wheel. Steering angle displacement $\delta_i$ is measured deviation from the forward driving direction, with $\delta_i=\angle\hat{B}_x\hat{S}_{xi}$ for $i = 1,4$ and $\delta_i=\angle\hat{B}_x\hat{S}_{xi}-\pi$ for $i = 2,3$. Unit Vectors $\hat{X}$, $\hat{Y}$, and $\hat{Z}$ are fixed in the Newtonian frame $O$. 

\subsection{Kinematics Parametrization, Singularity Analysis, and its Affect on Aerial Orientation Manipulability}
While airborne, we command symmetric movement of the front right and rear left wheels $(\delta_1 = \delta_3)$ and the front left and rear right wheels $(\delta_2 = \delta_4)$. This allows us to consider only two axes about which wheel reaction torque may be generated instead of four. We then divide the steering motion of these wheel pairs into two main coordinated submovements: one submovement $\alpha$ during which the two pairs of wheels move in opposing directions,
\begin{equation}
    \Delta \alpha=\Delta\delta_1 = -\Delta\delta_2
\end{equation}
and another submovement $\beta$ during which the two pairs of wheels move in the same direction.
\begin{equation}
    \Delta \beta=\Delta\delta_1 = \Delta\delta_2    
\end{equation}
Let us define the first opposing-direction submovement $\alpha$ as 
\begin{equation}
    \alpha = \frac{(\delta_1 +\delta_3) - (\delta_2 + \delta_4)}{4}
\end{equation}
and the second same-direction submovement $\beta$ as
\begin{equation}
    \beta = \frac{(\delta_1 + \delta_3) + (\delta_2 + \delta_4)}{4}
\end{equation}
where
\begin{equation}\begin{split}
    \delta_1 = \delta_3 = \beta + \alpha\\
    \delta_2 = \delta_4 = \beta - \alpha
\end{split}
\end{equation}

\begin{figure}[t]
	\centering
	\includegraphics[width=1\linewidth]{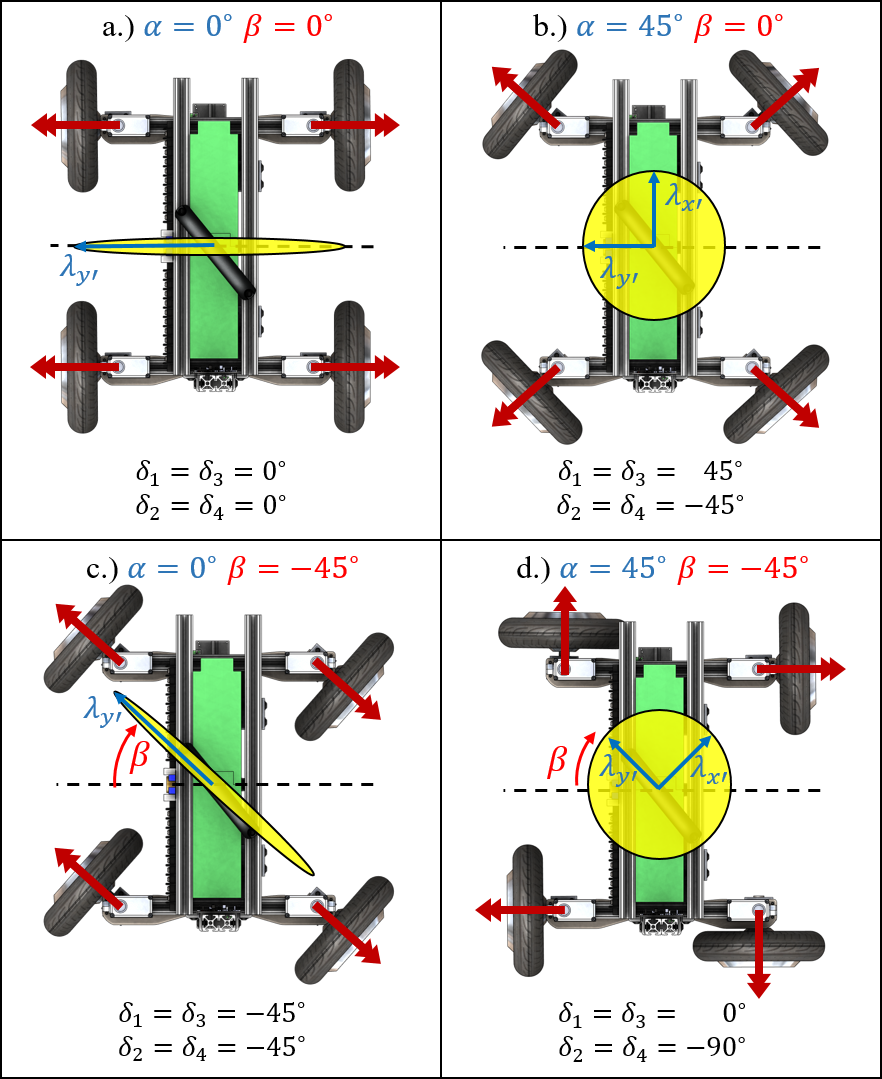}
	\caption{Kinematic parameters in the $\hat{B}_{xy}$ plane and their affect on the manipulability of orientation. Parameter $\alpha$ dictates the ratio between authority over $\hat{B}_{x}'$ axis rotation and $\hat{B}_{y}'$ axis rotation, with the manipulability ellispoid (yellow) collapsing at singular values.}
	\label{Kinematic Params XY}
\end{figure}

To visualize these submovements, consider the configurations shown in Fig. \ref{Kinematic Params XY}. Applying the symmetric submovement $\alpha = \frac{\pi}{4}$ to the neutral configuration shown in Fig. \ref{Kinematic Params XY}-a leads to the configuration shown in Fig. \ref{Kinematic Params XY}-b. Applying the asymmetric submovement $\beta = -\frac{\pi}{4}$ to the neutral configuration shown in Fig. \ref{Kinematic Params XY}-a leads to the configuration shown in Fig. \ref{Kinematic Params XY}-c. The superposition of these two movements leads to the configuration shown in Fig. \ref{Kinematic Params XY}-d. 

The coordinated submovements $\alpha$ and $\beta$ affect the dynamics of the robot travelling through the air in different ways. First, coordinated motion of the wheels along submovement $\alpha$ leads to zero net torque applied to the base along the $\hat{B}_z$ axis, and no yaw motion is observed. On the other hand, coordinated motion of the wheels along submovement $\beta$ lead to a net torque applied to the base along the $\hat{B}_z$ axis, and yaw motion of the base opposite the direction of wheel motion is observed in accordance with conservation of angular momentum. If $\alpha=\frac{\pi}{4}$, adjusting $\beta$ can be used to control the base yaw orientation with minimal effect on the roll and pitch or the ratio of roll and pitch authority.

The second effect of coordinated submovements $\alpha$ and $\beta$ on the airborne robot is to adjust the ratio and direction of orthogonal attitude control authority. Let the net roll torque on the base from the wheels about $\hat{B}_{x}$ be $\tau_x$ and net pitching torque on the base about $\hat{B}_{y}$ be $\tau_y$. The net torque about the first control authority axis $\hat{B}_{x}'$ is $\tau_x'$ and net torque about the second control authority axis $\hat{B}_{y}'$ is $\tau_y'$. The angle of $\hat{B}_{x}'$ relative to $\hat{B}_{x}$ is $\beta$. This leads to the following expressions for net base torque.
\begin{equation}\label{taux}
    \tau_x = \tau_x'\cos{\beta}-\tau_y'\sin{\beta}
\end{equation}
\begin{equation}\label{tauy}
    \tau_y = \tau_y'\cos{\beta}+\tau_x'\sin{\beta}
\end{equation}
If $\beta=0$, these orthogonal $\hat{B}_{x}'$ and $\hat{B}_{y}'$ axes lie along the base $\hat{B}_{x}$ and $\hat{B}_{y}$ axes, and are the roll and pitch of the base, respectively. Adjusting $\beta$ rotates the $\hat{B}_{x}'$ and $\hat{B}_{y}'$ axes about which the orientation control authority ratio may be adjusted.

Assuming $\alpha$ and $\beta$ are constant (and thus no forces result from inertial effects on the wheel), the magnitudes of $\tau_x'$ and $\tau_y'$ are determined by $\alpha$ and the individual wheel torques $\tau_{i}$ about $\hat{W}_{yi}$.
\begin{equation}\label{tauxprime}
    \tau_x' = \left(-\tau_1+\tau_2+\tau_3-\tau_4\right)\sin{\alpha}
\end{equation}
\begin{equation}\label{tauyprime}
    \tau_y' = \left(\tau_1+\tau_2-\tau_3-\tau_4\right)\cos{\alpha}
\end{equation}

Assuming symmetric application of torque ($\tau_1 = -\tau_3$ and $\tau_2 = -\tau_4$) and combining \eqref{taux}, \eqref{tauy}, \eqref{tauxprime}, and \eqref{tauyprime} leads to
\begin{equation}
    \begin{bmatrix}\tau_x\\ \tau_y\end{bmatrix} = 
    \mathbb{J}_{\tau'}\begin{bmatrix}\tau_1\\ \tau_2\end{bmatrix}
\end{equation}
where the configuration-dependent Jacobian $\mathbb{J}_{\tau'}$ between individual wheel torques and base pitch and roll torque is
\begin{equation}
    \mathbb{J}_{\tau'}= 
    \begin{bmatrix}
    2\sin(\alpha+\beta) & 2\sin(\alpha-\beta)\\
    -2\cos(\alpha+\beta) & 2\cos(\alpha-\beta)
    \end{bmatrix}
\end{equation}

The eigenvalues $\lambda_x'$ and $\lambda_y'$ of $\mathbb{J}_{\tau'}$ correspond with the authority over rotation about the $\hat{B}_{x}'$ axis and $\hat{B}_{y}'$ axis, respectively. While the isotropic configuration of [$\alpha=\frac{\pi}{4},~\beta=0$] is shown in Fig. \ref{Kinematic Params XY}-b, parameter $\alpha$ may be adjusted to provide more control authority over either the $\hat{B}_{x}'$ or $\hat{B}_{y}'$ axis. The determinant of $\mathbb{J}_\tau'$
\begin{equation}
    \det \mathbb{J}_\tau' = \sin(2\alpha)
\end{equation}
provides insight into singular configurations in which rotational authority about either the $\hat{B}_{x}'$ or $\hat{B}_{y}'$ axis becomes 0. Parameter $\alpha$ alone dictates relative control authority, with $\alpha = \pm \pi/4$ being isotropic configurations in which $\lambda_x'=\lambda_y'$, and $\alpha=0,\pm\pi/2$ being singular configurations in which $\lambda_x'$ or $\lambda_y'$ become 0. The Jacobian eigenvalue ellipsoid (manipulability ellipsoid) collapses along one axis at these singular configurations. See Fig. \ref{Kinematic Params XY} for a visualization. For example, [$\alpha=0,~\beta=0$] (wheels driving forward/back) provides full authority over the pitch and none over the roll, while [$\alpha=\frac{\pi}{2},~\beta=0$] (wheels driving left/right) provides full authority over roll and none over the pitch. 

Incorporating the net steering torques of $\beta$ motion by assuming all steering joints apply the same torque $\tau_{\delta}$ gives the full expression
\begin{equation}\label{torqueMatrix}
    \begin{bmatrix}\tau_x\\ \tau_y \\ \tau_z\end{bmatrix} = 
    \mathbb{J}_\tau\begin{bmatrix}\tau_1\\ \tau_2 \\ \tau_\delta\end{bmatrix}
\end{equation}
where the configuration-dependent Jacobian $\mathbb{J}_\tau$ between individual wheel and steering torques and the instantaneous base roll, pitch, and yaw torques is
\begin{equation}\label{Jacobian}
    \mathbb{J}_\tau= 
    \begin{bmatrix}
    2\sin(\alpha+\beta) & 2\sin(\alpha-\beta) & 0\\
    -2\cos(\alpha+\beta) & 2\cos(\alpha-\beta) & 0\\
    0 & 0 & 4
    \end{bmatrix}
\end{equation}

By combining coordinated motion $\alpha$ and $\beta$ with the reaction torque from spinning the wheels, the orientation of AGRO while in the air can be controlled, assuming no steering angle or wheel velocity limits are reached.

\subsection{Dynamics of an Airborne 4WIDS Robot}
Before designing the base orientation controller, the dynamic equations of the system must be obtained. These dynamics will enable the system to be simulated and the controller to be designed and tested before being implemented on the prototype.

\begin{figure}[ht]
	\centering
	\includegraphics[width=1\linewidth]{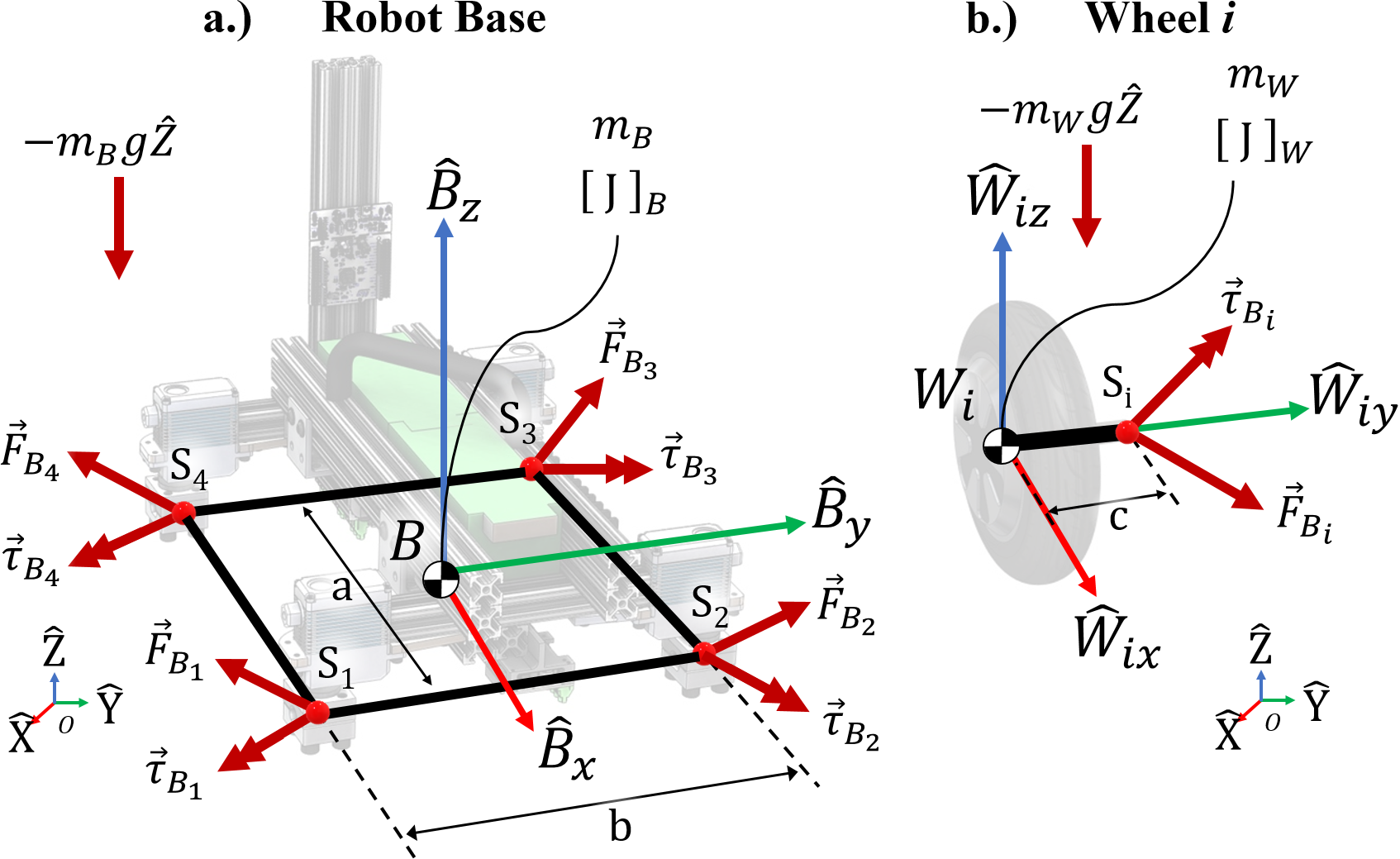}
	\caption{Free Body Diagram for a). the main base and b). a single wheel module.}
	\label{FBD}
\end{figure}
Gravity acts on the base of the robot at $B$ and the four attached steerable offset wheels impart reaction torques $\Vec{\tau}_{Bi}$ and forces $\Vec{F}_{Bi}$ on the base at points $S_i$ (See Fig. \ref{FBD}). Taking the Newton-Euler equation for 
angular momentum for the base about its mass center, we get
\begin{equation}
    \sum_{i=1}^{4}\left(\Vec{\tau}_{Bi}+\Vec{r}_{BS_i}\times\Vec{F}_{Bi}\right) =
    \frac{d}{dt}\left(\left[J_B\right]\Vec{\Omega}\right)
\end{equation}
where the angular momentum of the base is
\begin{equation}
    \left[J_B\right]\Vec{\Omega} = 
    \begin{cases}
    J_{Bxx}\Omega_x \hat{B}_x\\ 
    J_{Byy}\Omega_y \hat{B}_y\\
    J_{Bzz}\Omega_z \hat{B}_z
    \end{cases}
\end{equation}
The four steering axes are located symmetrically about $B$ as follows
\begin{equation}
    \begin{split}
    \Vec{r}_{BS_1} &= \frac{a}{2}\hat{B}_x - \frac{b}{2}\hat{B}_y \\
    \Vec{r}_{BS_2} &=  \frac{a}{2}\hat{B}_x  +\frac{b}{2}\hat{B}_y\\
    \Vec{r}_{BS_3} &=  -\frac{a}{2}\hat{B}_x  +\frac{b}{2}\hat{B}_y \\
    \Vec{r}_{BS_4} &= -\frac{a}{2}\hat{B}_x - \frac{b}{2} \hat{B}_y
    \end{split}
\end{equation}
leading to the following angular momentum equation for the base:
\begin{equation}\label{bodyRot1}
    \begin{split}
    &\frac{d}{dt}\left(\left[J_B\right]\Vec{\Omega}\right) = \Vec{\tau}_{B1}+\Vec{\tau}_{B2}+\Vec{\tau}_{B3}+\Vec{\tau}_{B4}\\
    &+\Vec{r}_{BS_1}\times(\Vec{F}_{B1} -\Vec{F}_{B3})+ \Vec{r}_{BS_2}\times(\Vec{F}_{B2} -\Vec{F}_{B4})
    \end{split}
\end{equation}

Resultant reaction forces $\Vec{F}_{Bi}$ can be solved for using the linear momentum equation for the wheels.
\begin{equation}\label{FBI}
    \Vec{F}_{Bi} =
    -m_W\left(\frac{d^2}{dt^2}\left(\Vec{r}_{OW_i}\right)+g\hat{Z}\right)
\end{equation}
For tractability, we assume a nonmoving steering angle ($\delta$ constant, $\dot\delta=\ddot\delta=0$). The displacement, velocity, and acceleration of the wheel in the Newtonian frame are then
\begin{equation}
    \Vec{r}_{OW_i} = \Vec{r}_{OB} + \Vec{r}_{BS_i} + \Vec{r}_{S_iW_i}
\end{equation}
\begin{equation}
    \frac{d}{dt}\Vec{r}_{OW_i}=\Vec{v}_{OW_i} = \Vec{v}_{OB} + \Vec{\Omega}\times\left(\Vec{r}_{BS_i}+\Vec{r}_{S_iW_i}\right)
\end{equation}
\begin{equation}
\begin{split}
    \frac{d^2}{dt^2}\Vec{r}_{OW_i}=\Vec{a}_{OW_i} =& \Vec{a}_{OB} + \frac{d}{dt}\Vec{\Omega}\times\left(\Vec{r}_{BS_i}+\Vec{r}_{S_iW_i}\right)\\ 
    &+\Vec{\Omega}\times\left(\Vec{\Omega}\times\left(\Vec{r}_{BS_i}+\Vec{r}_{S_iW_i}\right)\right)
\end{split}
\end{equation}
where
\begin{equation}
    \frac{d}{dt}\Vec{\Omega} =  \dot{\Vec{\Omega}}= 
    \begin{cases}
    \dot\Omega_x \hat{B}_x\\ 
    \dot\Omega_y \hat{B}_y\\
    \dot\Omega_z \hat{B}_z
    \end{cases}
\end{equation}
and
\begin{equation}
    \Vec{r}_{S_iW_i} = -\Vec{r}_{W_iS_i}= -c\hat{W_i}_y
\end{equation}

Combining \eqref{bodyRot1} and \eqref{FBI} gives
\begin{equation}\
    \begin{split}
    \frac{d}{dt}\left(\left[J_B\right]\Vec{\Omega}\right)
    =&\Vec{\tau}_{B1}+\Vec{\tau}_{B2}+\Vec{\tau}_{B3}+\Vec{\tau}_{B4}\\
    -2m_W\Vec{r}_{BS_1}\times(&\dot{\Vec{\Omega}}\times(\Vec{r}_{BS_1}+\Vec{r}_{S_1W_1})\\
    &+\Vec{\Omega}\times(\Vec{\Omega}\times(\Vec{r}_{BS_1}+\Vec{r}_{S_1W_1})))\\
    -2m_W\Vec{r}_{BS_2}\times(&\dot{\Vec{\Omega}}\times(\Vec{r}_{BS_2}+\Vec{r}_{S_2W_2})\\    
    &+\Vec{\Omega}\times(\Vec{\Omega}\times(\Vec{r}_{BS_2}+\Vec{r}_{S_2W_2})))
    \end{split}
\end{equation}
Note that moments due to gravity and base acceleration cancel out due to symmetry. 

Torque on the wheel $\Vec{\tau}_{Bi}$ can be broken down as the wheel torque input $\tau_i$, the steering joint torque $\tau_{\delta i}$ and one reaction torque $\tau_x$. 
\begin{equation}\label{torques}
    \Vec{\tau}_{Bi} = 
    \begin{cases}
    \tau_{xi} \hat{W_i}_x \\
    \tau_i \hat{W_i}_y \\
    \tau_{\delta i} \hat{W_i}_z    
    \end{cases}
\end{equation}

Because $\delta_1 = \delta_3$, $\tau_1 = -\tau_3$, $\delta_2 = \delta_4$, $\tau_2 = -\tau_4$, and assuming all $\tau_{\delta i}$ are equal we can simplify using \eqref{torqueMatrix}
\begin{equation}\label{eqPretty}
\begin{split}
    &\frac{d}{dt}\left(\left(\left[J_B\right]+\left[J_{m_W}\right]\right)\Vec{\Omega}\right) =\mathbb{J}_\tau\begin{bmatrix}\tau_1\\ \tau_2 \\ \tau_\delta\end{bmatrix}
    +\mathbb{J}_x\begin{bmatrix}\tau_{x1} \\ \tau_{x2} \\ \tau_{x3} \\ \tau_{x4}\end{bmatrix}
\end{split}
\end{equation}
where the reflection of wheel masses to the overall inertia of the base is
\begin{equation}\begin{split}
    & \left[J_{m_W}\right]=\\
    &\begin{bmatrix}
    J_{m_Wxx} & 0 & 0 \\
    0 & J_{m_Wyy}& 0 \\
    0 & 0 & J_{m_Wxx} + J_{m_Wyy}
    \end{bmatrix}
\end{split}
\end{equation}
where
\begin{equation}
    J_{m_Wxx} = 2m_W\left((\frac{b}{2} + c\cos\delta_1)^2+(\frac{b}{2} + c\cos\delta_2)^2\right)
\end{equation}
and 
\begin{equation}
    J_{m_Wyy} = 2m_W\left((\frac{a}{2} + c\sin\delta_1)^2+(\frac{b}{2} + c\sin\delta_2)^2\right)
\end{equation}
and the matrix converting wheel reaction torques $\tau_{xi}$ to the $\hat{B}_{xyz}$ frame is
\begin{equation}\label{Jacobian2}
    \mathbb{J}_x= 
    \begin{bmatrix}
    \cos\delta_1 & -\cos\delta_2 & -\cos\delta_1 & \cos\delta_2\\
    \sin\delta_1 & -\sin\delta_2 & -\sin\delta_1 & \sin\delta_2\\
    0 & 0 & 0 & 0\\
    \end{bmatrix}
\end{equation}

The scalar wheel reaction torque component $\tau_{xi}$ can be found by taking the Newton-Euler equation for 
angular momentum for a single general wheel $i$ about its center of mass
\begin{equation}\label{rotEq2-1} 
    -\Vec{\tau}_{Bi}-\Vec{r}_{W_iS_i}\times\Vec{F}_{Bi} =\frac{d}{dt}\left(\left[J_W\right]\Vec{\omega}_i\right)
\end{equation}
and dotting with $\hat{W}_{ix}$
\begin{equation}\label{Tauxi}
    \tau_{xi}=\left(-\frac{d}{dt}\left(\left[J_W\right]\Vec{\omega}_i\right)-\Vec{r}_{W_iS_i}\times\Vec{F}_{Bi}\right) \cdot \hat{W}_{ix}
\end{equation}

Combining equations \eqref{eqPretty} through \eqref{Tauxi} yields scalar equations of motion for the base in terms of the reaction forces and torques.
For tractability, these equations assume that gravity is aligned in the $\hat{B}_z$ and again that the derivatives of $\delta_i$ are zero. 

Equation of motion about the $\hat{B}_x$ axis:
\begin{equation} \label{EOMx}
\begin{split}
(J_{Bxx}+J_{m_Wxx}+2J_{Wxx}(\cos\delta_1+\cos\delta_2))\dot{\Omega}_x&\\
+(J_{Bzz}-J_{Byy} + J_{m_Wxx})\Omega_y\Omega_z&\\
= -2\tau_1 \cos\delta_1+2\tau_2\cos&\delta_2
\end{split}
\end{equation}

Equation of motion about the $\hat{B}_y$ axis:
\begin{equation}\label{EOMy}
\begin{split}
(J_{Byy}+J_{m_Wyy}+2J_{Wxx}(\sin\delta_1+\sin\delta_2))\dot\Omega_y&\\
+(J_{Bxx}-J_{Bzz}+ J_{m_Wyy})\Omega_x\Omega_z&\\
=2\tau_1 \sin \delta_1+2\tau_2\sin&\delta_2
\end{split}
\end{equation}

Equation of motion about the $\hat{B}_z$ axis:
\begin{equation} \label{EOMz}
\begin{split}
(J_{Bzz})\dot\Omega_z + (J_{Byy}-J_{Bxx})\Omega_x\Omega_y&=4\tau_{\delta}
\end{split}
\end{equation}
where $J_{Wxx}$ is the component of $\left[J_W\right]$ along $\hat{W}_{ix}$.

\section{Analysis and Simulation of\\Aerial Attitude Control}\label{Control}
We now aim to control the orientation of the robot base $\Vec{q}$ using the wheel torques $\tau_i$ and net steering torque $\tau_{\delta}$. During operation, AGRO automatically detects freefall and enables the attitude controller. To do this, AGRO runs a finite state machine that gives the operator command over the vehicle translation and rotation when on the ground. The on-board accelerometer detects the gravitational acceleration vector of magnitude $1~g$ ($9.8~m/s^2$). When the accelerometer detects no acceleration, it considers itself to be in freefall, takes command away from the user, and changes to the free-falling orientation control state.

While in the freefall state, AGRO immediately sets $\alpha = 45^\circ$ (shown in Fig. \ref{Kinematic Params XY}-b) to equalize authority over both pitch and roll. While $\alpha$ can be tuned proportionally to the demand of pitch vs roll, we do not implement this functionality in this work. In this configuration, AGRO can use its wheel torques $\tau_i$ and net steering torques $\tau_{\delta i}$ in coordination to generate net moments about its base and regulate its orientation. 

Let the small deviation of the base orientation $\Vec{q}$ from desired equilibrium $\Vec{q}_{desired}$ be
\begin{equation}
    \Vec{q} = \begin{bmatrix}\varphi& \theta & \psi \end{bmatrix}^T
\end{equation}
Equations \eqref{EOMx}, \eqref{EOMy}, and \eqref{EOMz} can be linearized about the equilibrium $\delta_1=\pi/4$, $\delta_2=-\pi/4$, $\Omega_x = \Omega_y = \Omega_z = 0$ to produce decoupled linear equations of motion for control design:
\begin{equation} \label{EOMxLin}
\begin{split}
\left(J_{Bxx}+J_{m_Wxx}+2\sqrt{2}J_{Wxx}\right)\dot{\Omega}_x
= \sqrt{2}(\tau_2 - \tau_1)
\end{split}
\end{equation}
\begin{equation}\label{EOMyLin}
\begin{split}
\left(J_{Byy}+J_{m_Wyy}+2\sqrt{2}J_{Wxx}\right)\dot\Omega_y
=\sqrt{2}(\tau_1+\tau_2)
\end{split}
\end{equation}
\begin{equation} \label{EOMzLin}
J_{Bzz}\dot\Omega_z
=4\tau_{\delta}
\end{equation}
which, when combined with \eqref{torqueMatrix}, give transfer functions in terms of $\Vec{q}$
\begin{equation} \label{TFx}
\frac{\varphi}{\tau_x} (s) = 
\frac{1}{(J_{Bxx}+J_{m_Wxx}+2\sqrt{2}J_{Wxx})s^2}
\end{equation}
\begin{equation} \label{TFy}
\frac{\theta}{\tau_y} (s) = 
\frac{1}{\left(J_{Byy}+J_{m_Wyy}+2\sqrt{2}J_{Wxx}\right)s^2}
\end{equation}
\begin{equation} \label{TFz}
\frac{\psi}{\tau_z} (s) = 
\frac{1}{(J_{Bzz})s^2}
\end{equation}

These systems can be stabilized with simple PD controllers using the following feedback control law
\begin{equation}
    \vec{\tau}_{B}=\begin{bmatrix}\tau_x\\ \tau_y \\ \tau_z\end{bmatrix} =K_P\left(\Vec{q}_{desired} - \Vec{q}\right) - K_D\dot{\Vec{q}}
\end{equation}
where $K_P$ and $K_D$ are the symmetric positive definite proportional and derivative gain matrices. Because we consider the linear decoupled systems, these matrices are also diagonal. 
\begin{figure}[t]
	\centering
	\includegraphics[width=1\linewidth]{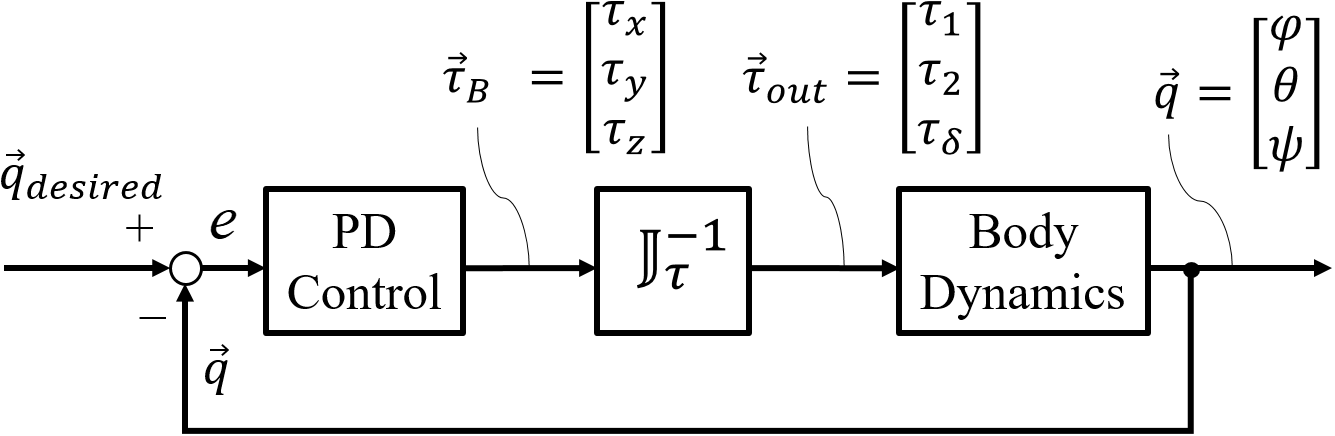}
	\caption{Block Diagram of the Aerial Attitude Controller.}
	\label{Control Diagram}
\end{figure}

The commanded wheel torques $\tau_i$ and steering torques $\tau_{\delta i}$ can be derived from the torques $\tau_x$, $\tau_y$, and $\tau_z$ in the $B$ frame by inverting \eqref{torqueMatrix}, assuming the robot does not take a singular configuration, leading to the following control law\begin{equation}
    \vec{\tau}_{out}=\begin{bmatrix}\tau_1\\ \tau_2 \\ \tau_\delta\end{bmatrix} =\mathbb{J}_\tau^{-1} K_P\left(\Vec{q}_{desired} - \Vec{q}\right) - \mathbb{J}_\tau^{-1}K_D\dot{\Vec{q}}
\end{equation}
Note that we are assuming symmetric application of torque ($\tau_1 = -\tau_3$ and $\tau_2 = -\tau_4$) and that all wheels apply equal steering torque $\tau_{\delta}$. See Fig. \ref{Control Diagram} for a block diagram of the entire control system.
\begin{figure}[ht]
	\centering
	\includegraphics[width=0.875\linewidth]{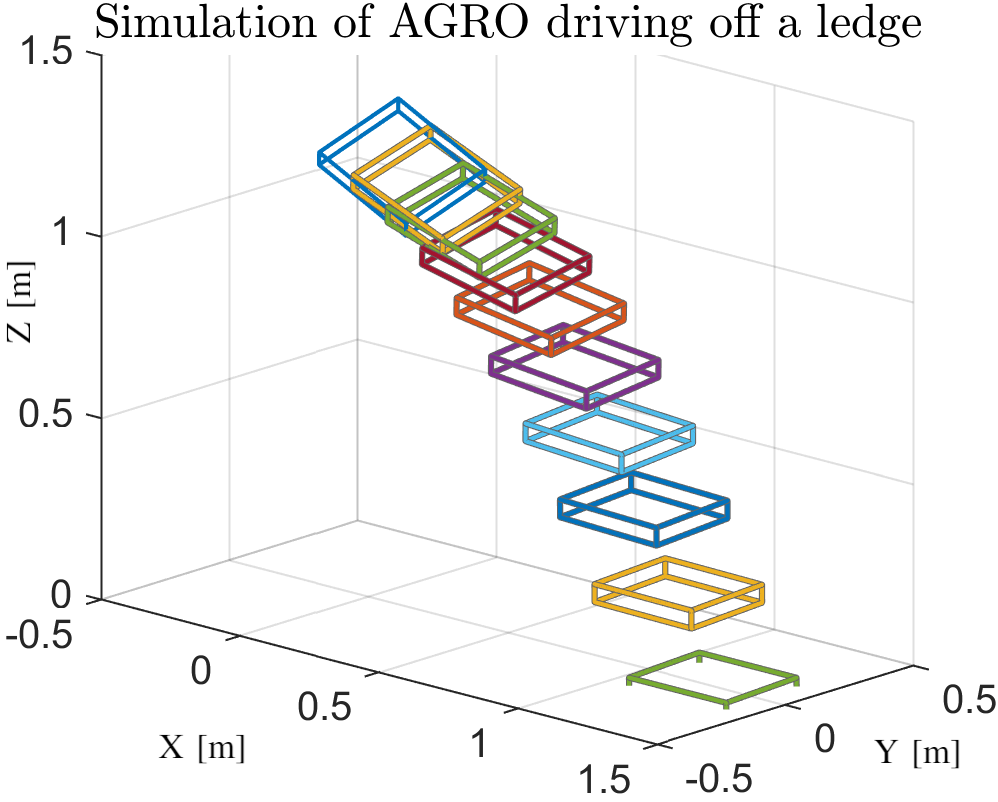}
	\caption{Simulation of Robot Aerial Attitude Control Response to Initial Conditions.}
	\label{ControlSim1}
\end{figure}

To tune the controller, the nonlinear dynamics of AGRO falling and reacting to the wheel torques were simulated using an ODE solver. The mass and inertia properties in the simulated robot match those of the AGRO prototype.

\begin{figure}[t]
	\centering
	\includegraphics[width=1\linewidth]{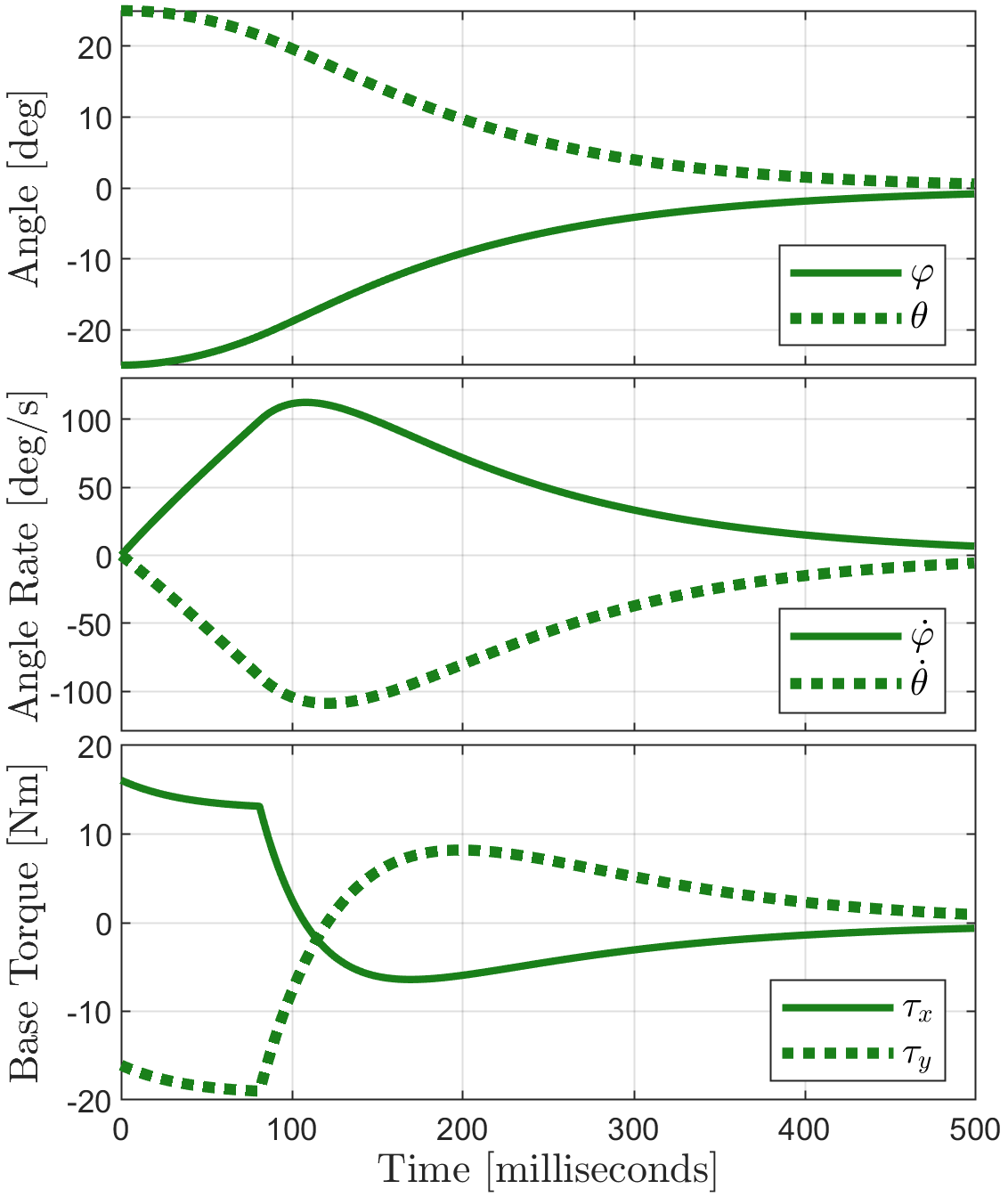}
	\caption{Base Roll and Pitch angles, angular velocities, and commanded base torques $\tau_x$ and $\tau_y$ during a simulation of AGRO driving off a ledge.}
	\label{ControlSim2}
\end{figure}

A test case shown in shown in Fig. \ref{ControlSim1} and  Fig. \ref{ControlSim2} involves the robot driving off of a 1.2 meter ledge at a forward velocity of 2.5 m/s and given an angle disturbance in the form of an initial angle condition of $\phi = -25^\circ$ and $\theta = 25^\circ$. The yaw controller was disabled by tuning its proportional and derivative gain parameters to 0, and the pitch and roll controllers each have a proportional gain of 75 Nm/rad and a derivative gain of 12 Nms/rad. The simulated robot reaches the desired zero angle before impact at 498 milliseconds.

\section{Experimental Validation of Aerial Attitude Control}\label{Experiment}
\subsection{Experiment Setup}
\begin{figure}[ht]
	\centering
	\includegraphics[width=1\linewidth]{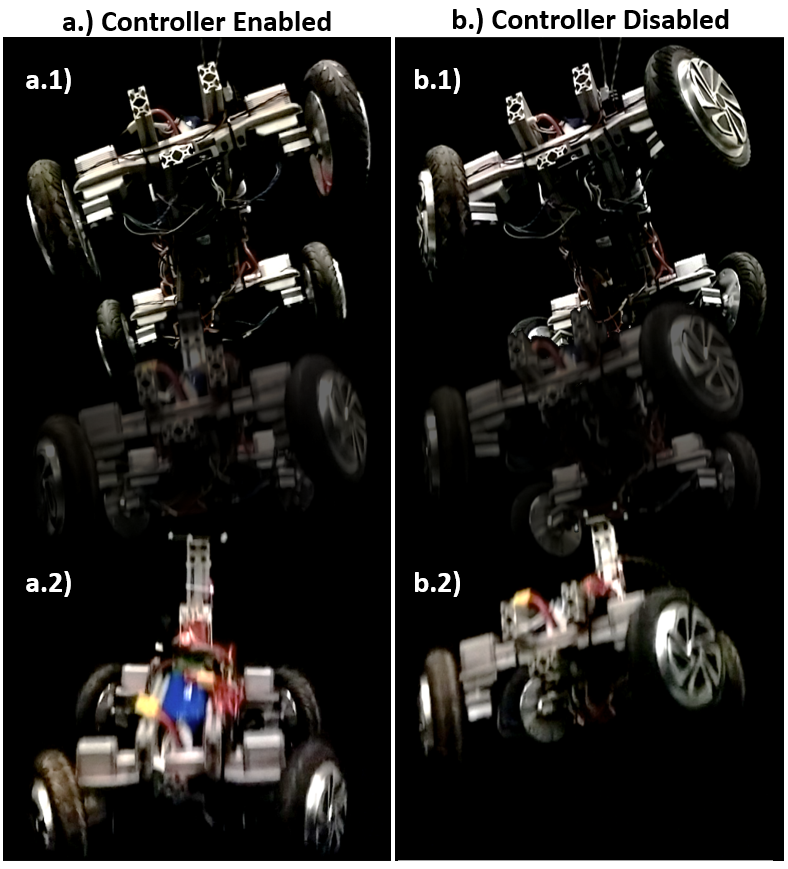}
	\caption{High-speed video stills of AGRO being dropped from 0.85 m with the attitude controller a.) enabled and b.) disabled. Each trial was started at the same initial condition at frames a.1) and b.1) and touched down on the landing pad at frames a.2) and b.2). \textbf{Experiment video available in supplemental materials.}}
	\label{ExpSetup}
\end{figure}
An experiment was designed to evaluate the roll and pitch control of AGRO as it fell to the ground. A comparison was made between a fall utilizing the controller to stabilize the base and a fall with no control enabled. As shown in Fig. \ref{ExpSetup}-a.1 and Fig. \ref{ExpSetup}-b.1, the AGRO prototype was suspended from a tether on a cantilever beam with initial conditions of roll, pitch, and free-fall height. Because AGRO currently lacks a suspension to mitigate impact, a foam landing pad was placed underneath. The robot tether was released from the end of the cantilever beam to begin freefall.

The free-fall height was set to $h=0.85$ m to prevent damage to the robot during the uncontrolled fall while providing the controlled fall 0.416 seconds to stabilize before impact. The initial roll and pitch angles were $16^\circ$ and $-23^\circ$, respectively. Prior to each trial, the robot was configured with the same initial conditions. 

For this experiment, the yaw controller was disabled by tuning its proportional and derivative gain parameters to 0. The pitch and roll controllers each have a proportional gain of 75 Nm/rad and a derivative gain of 12 Nms/rad, just like the simulation. 
\subsection{Results and Discussion}
\begin{figure}[ht]
	\centering
	\includegraphics[width=1\linewidth]{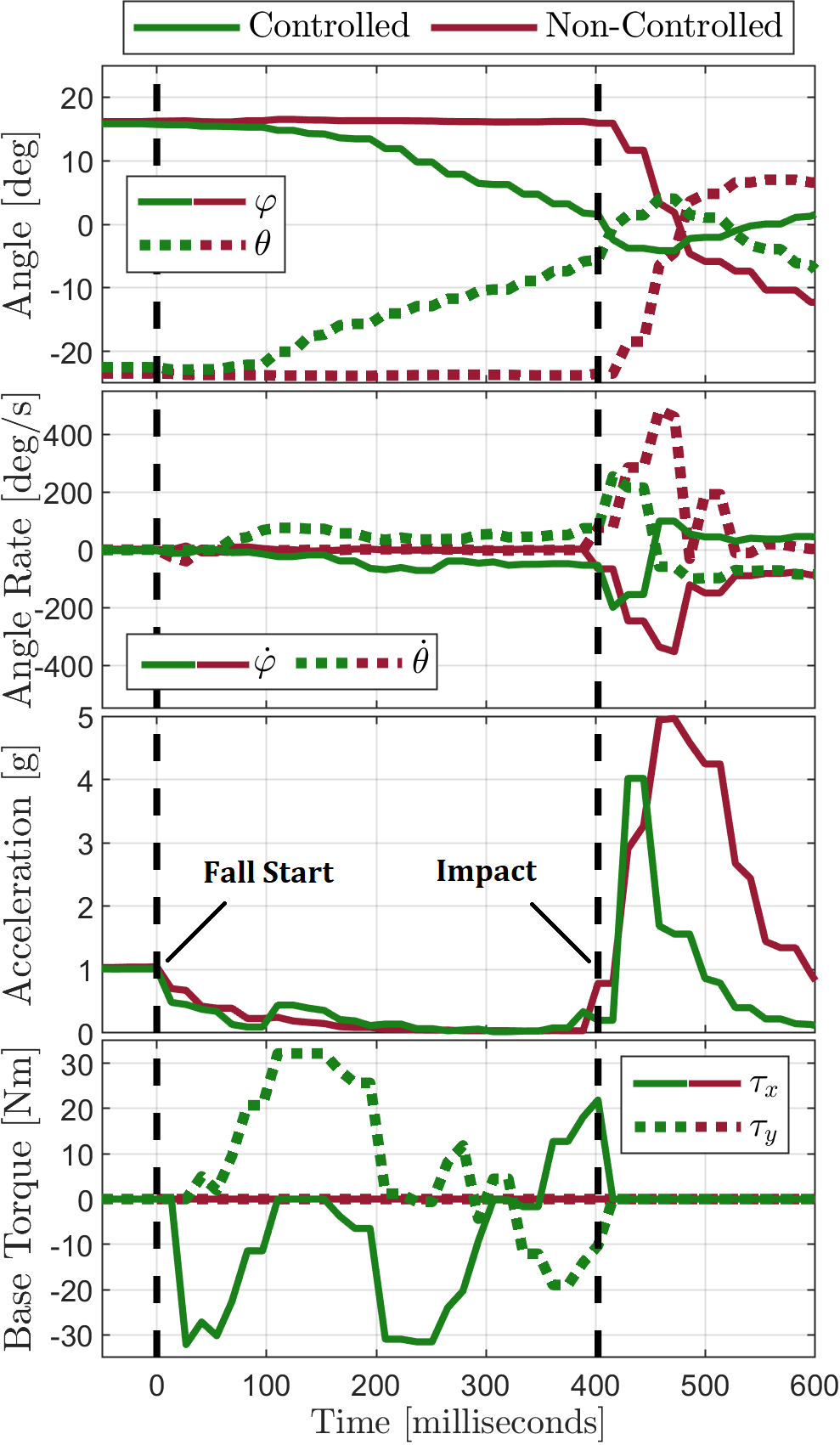}
	\caption{Measured roll and pitch angles, angle velocity, base acceleration, and commanded base torque during a controlled (green) and non-controlled (red) fall from initial conditions.}
	\label{Results 1}
\end{figure}
Fig. \ref{Results 1} shows the results of the experiment comparing a controlled (green) fall and a non-controlled (red) fall. The controlled and non-controlled robots have nearly identical initial pitch and roll angles before freefall ($t<0$) and when the fall starts ($t = 0$). 

At the time before impact ($t=402$ milliseconds), the controlled (green) roll angle is within $10\%$ of its initial condition and reached $-1.57^\circ$ and the controlled (green) pitch angle has reached $-0.48^\circ$. The non-controlled pitch and roll angles and angle rates (red) have not changed from their initial condition by the time of impact. See \ref{ExpSetup}-a.2 and \ref{ExpSetup}-b.2 to visualize the difference in orientation at impact.

After impacting the landing pad, the robot with attitude control disabled (red) lands on its rear-right wheel (Wheel 3), which causes an impact acceleration of 5 g on the body. This causes the impacting wheel to bounce back up, causing the whole robot to pitch forward and roll left before finally settling. On the other hand, the robot with attitude control enabled touches down on all four wheels simultaneously and only sees an acceleration of 4 g on its body, a 1 g improvement over the non-controlled case. Because the robot has angular velocity in the direction of overall rotation, the impact also imparts motion, causing a backward pitch and rightward roll before settling. 

The robot lands on all four wheels during the controlled fall, but has a nonzero angular velocity, which imparts uneven force between the wheels, albeit not as significant as the force of landing on one wheel. During the first half of the fall, motors 2 and 4 are commanded to their maximum torque limits. Higher torque limits and control gains would decrease the settling time and allow the controlled robot to stabilize its orientation and angular velocity before impact when dropped from this height.

\section{Conclusion and Future Work}\label{Conclusion}
In this work we analyzed and demonstrated aerial attitude control using drive wheel reaction torques on AGRO, a Four-Wheeled Independent Drive and Steering (4WIDS) Agile Ground RObot which was designed for rapid deployment by emergency response personnel into unsafe areas by being thrown. By adjusting its steering angles, AGRO can adjust the ratio between pitch and roll authority. An experiment was conducted in which AGRO was dropped from a height of 0.85 meters from an initial angle offset both with and without the controller enabled. AGRO was able to stabilize its orientation, touch down on all four of its wheels, and reduce impact acceleration by 1 g. 

Future work includes increasing the wheel torque and speed limits by operating at a higher voltage, which will allow for a greater total reaction impulse to be applied to the robot by the wheels. We will implement and test a strategy for adjusting the steering parameters to maximize control authority over the more critical axis of rotation, and take into account the gyroscopic dynamics that will emerge from the steering. AGRO 2 will incorporate an independent legged suspension allowing it to absorb landing impact forces and hop up stairs and curbs.
\bibliographystyle{IEEEtran}
\bibliography{IEEEabrv,biblio}

\end{document}